\documentclass[conference]{IEEEtran}
\IEEEoverridecommandlockouts
\usepackage{cite}
\usepackage{amsmath,amssymb,amsfonts,gensymb}
\usepackage{mathrsfs}
\usepackage{algorithmic}
\usepackage{graphicx}
\usepackage{textcomp}
\usepackage{xcolor}
\usepackage{bm}
\usepackage{subfigure}
\usepackage{hyperref}
\usepackage{lmodern}
\usepackage{makecell, booktabs}
    
\begin{document}

\title{}
\title{Walnut Detection Through Deep Learning Enhanced by Multispectral Synthetic Images\\

\thanks{$^{1}$The author is with the Department of Electrical and Computer Engineering, University of California, Davis, Davis, CA 95616, USA}

\thanks{$^{2}$The authors are with the Department of Plant Sciences, University of California, Davis, Davis, CA 95616, USA}

\thanks{*these authors contributed equally}

}

\author{Kaiming Fu$^{*1}$, Tong Lei$^{*2}$, Maryia Halubok $^{2}$, and Brian N. Bailey$^{2}$
}

\maketitle
\begin{abstract}
The accurate identification of walnuts within orchards brings forth a plethora of advantages, profoundly amplifying the efficiency and productivity of walnut orchard management. Nevertheless, the unique characteristics of walnut trees, characterized by their closely resembling shapes, colors, and textures between the walnuts and leaves, present a formidable challenge in precisely distinguishing between them during the annotation process. In this study, we present a novel approach to improve walnut detection efficiency, utilizing YOLOv5 trained on an enriched image set that incorporates both real and synthetic RGB and NIR images. Our analysis comparing results from our original and augmented datasets shows clear improvements in detection when using the synthetic images.
\end{abstract}

\section{Extended Summary}
\subsection{Introduction}
The walnut industry plays a crucial role in agriculture, highlighting its significant importance. Accurate walnut detection enhances yield estimation, optimizing resource allocation and harvest scheduling. Nonetheless, the distinctive attributes of walnut trees, characterized by their similarity in shape, color, and texture between both walnuts and leaves, pose a complex challenge to accurately distinguish between them during annotation. To address this challenge, the integration of near-infrared (NIR) imagery emerges as a potential solution, surpassing the limitations of relying solely on RGB images \cite{lee2014hyperspectral}. NIR imaging reveals subtle plant health and physiological features outside the visible spectrum that are often undetectable in standard RGB images \cite{sa2016deepfruits}. However, the collection of both RGB and NIR images encounters constraints due to the specific lighting conditions, particularly concerning large-sized trees. Consequently, these limitations impede the effective deployment of computer vision methods for precise walnut detection and monitoring, impacting the potential efficiency of agricultural robotics and automation applications within walnut orchards.

The challenge of accurately distinguishing between walnuts and leaves in object detection using computer vision can be effectively addressed by integrating synthetic images \cite{sa2022deepnir}. Synthetic images offers the advantage of creating extensive and diverse image sets, overcoming limitations in collecting real-world images that can be time-consuming, expensive, or impractical. This abundance of images enhances the model's capacity to generalize across various scenarios, resulting in improved robustness and accuracy during real-world deployment. One innovative approach for synthetic image generation is the utilization of radiative transfer-based models, which simulate the interactions of light with objects and environments, generating physically accurate synthetic data for both RGB and NIR images with fully resolved reference labels such as plant structures and the chemical concentrations of leaves \cite{lei2023radiative}. This technique is particularly valuable in scenarios where lighting conditions and environmental factors significantly influence image appearance. 

\textit{Contributions: } To improve crop detection in walnut trees, we added synthetic RGB and NIR images to our dataset. This study compares results from our original image set with the enhanced set, showing clear advantages of using synthetic images over just real ones.

\subsection{Methodology}
Helios \cite{bailey2019helios} is a flexible framework designed to smoothly combine and run various 3D environmental system models. It lets users build and detail 3D plant and soil models with specific optical, biophysical, and chemical features. A key feature in Helios is the Canopy Generator plug-in, which provides an easy-to-use interface for creating various plant canopy shapes. These shapes are defined by data structures that contain key geometric parameters, making it easy to customize specific canopy attributes.

Synthetic walnut images are generated and labeled by using Helios Radiation Model plug-in. This plug-in calculates the distribution of absorbed, reflected, transmitted, and emitted radiation for scene objects in one scattering iteration using a reverse ray-tracing method \cite{RAYTRACYING}. A camera model based on ray-tracing then measures the reflected and transmitted energy for each camera pixel across all wavelengths. To determine the walnut fruit label for an image pixel, a ray is sent from the camera. The label of the closest shape hit by this ray is then identified.

\subsection{Preliminary Results}
\begin{figure}
	\centering
	\subfigure[]{
		\includegraphics[width=1.5in]{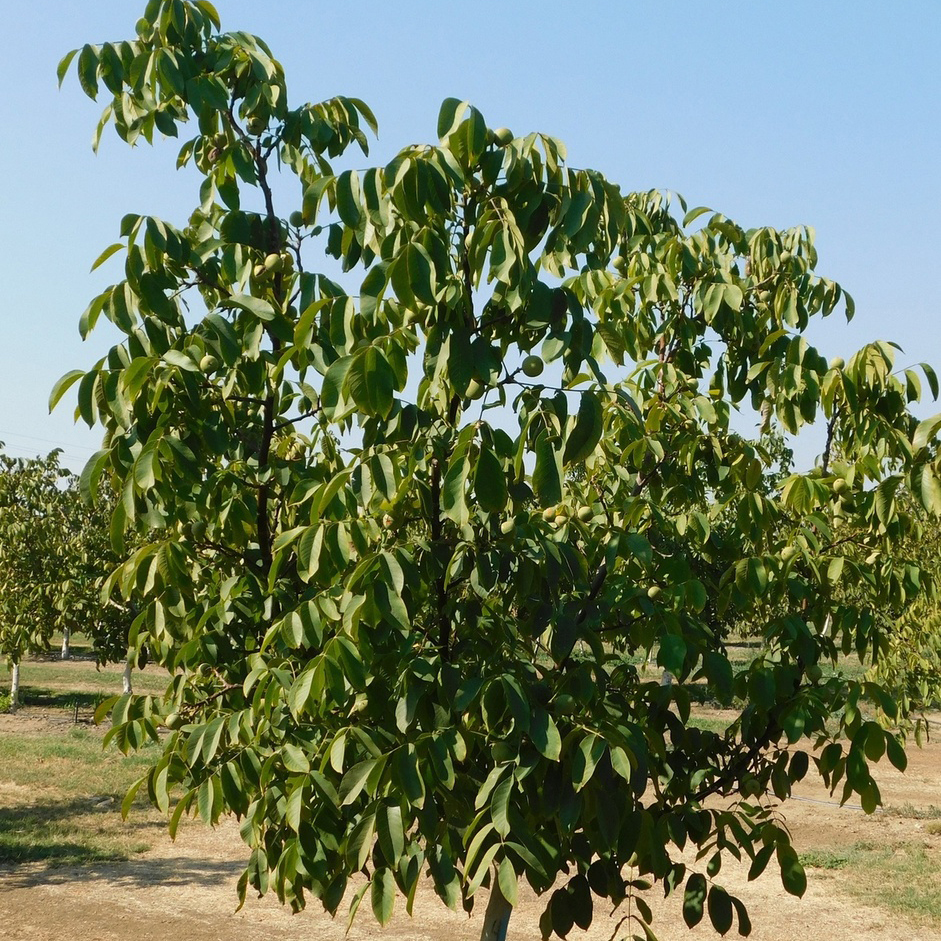}
	}
	\subfigure[]{
		\includegraphics[width=1.5in]{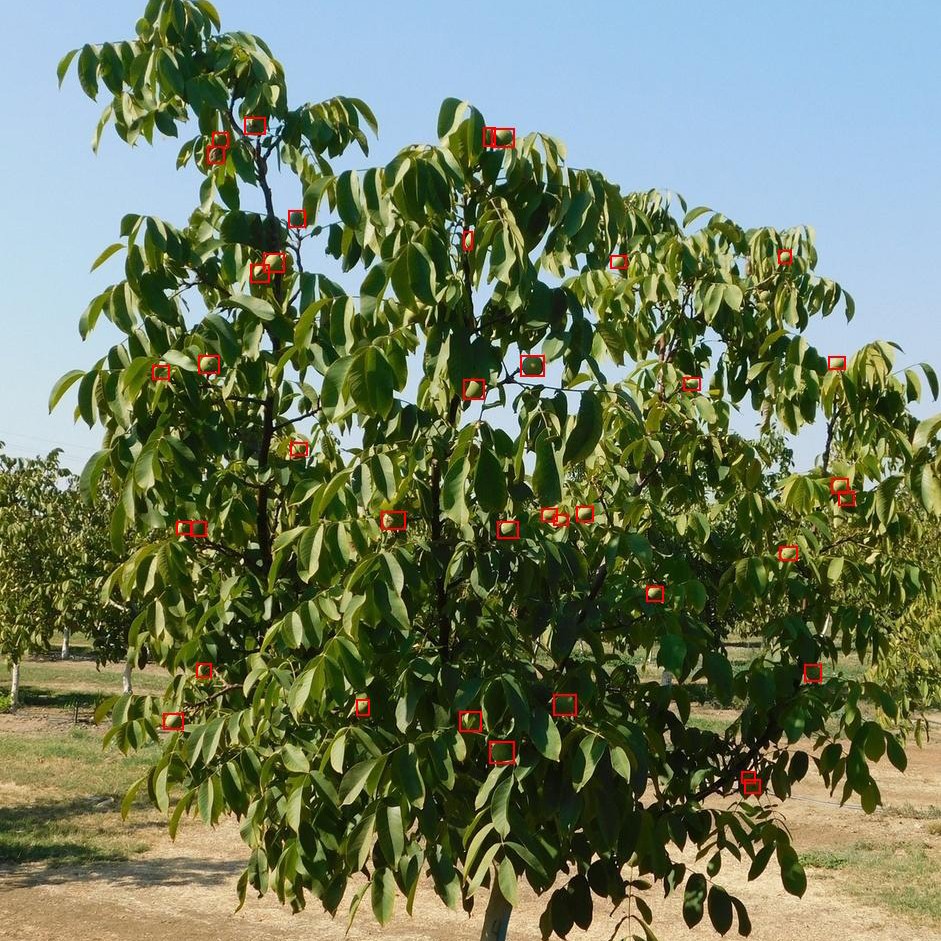}
	}	
	
	\caption{(a) RGB image captured with Nikon B500 camera, (b) Annotated RGB image. The camera was positioned 1m to 2m away from the canopy's region of interest (ROI).}
	\label{fig:Nikon_RGB_NIR}
\end{figure}

\begin{figure}
	\centering
	\subfigure[]{
		\includegraphics[width=1.5in]{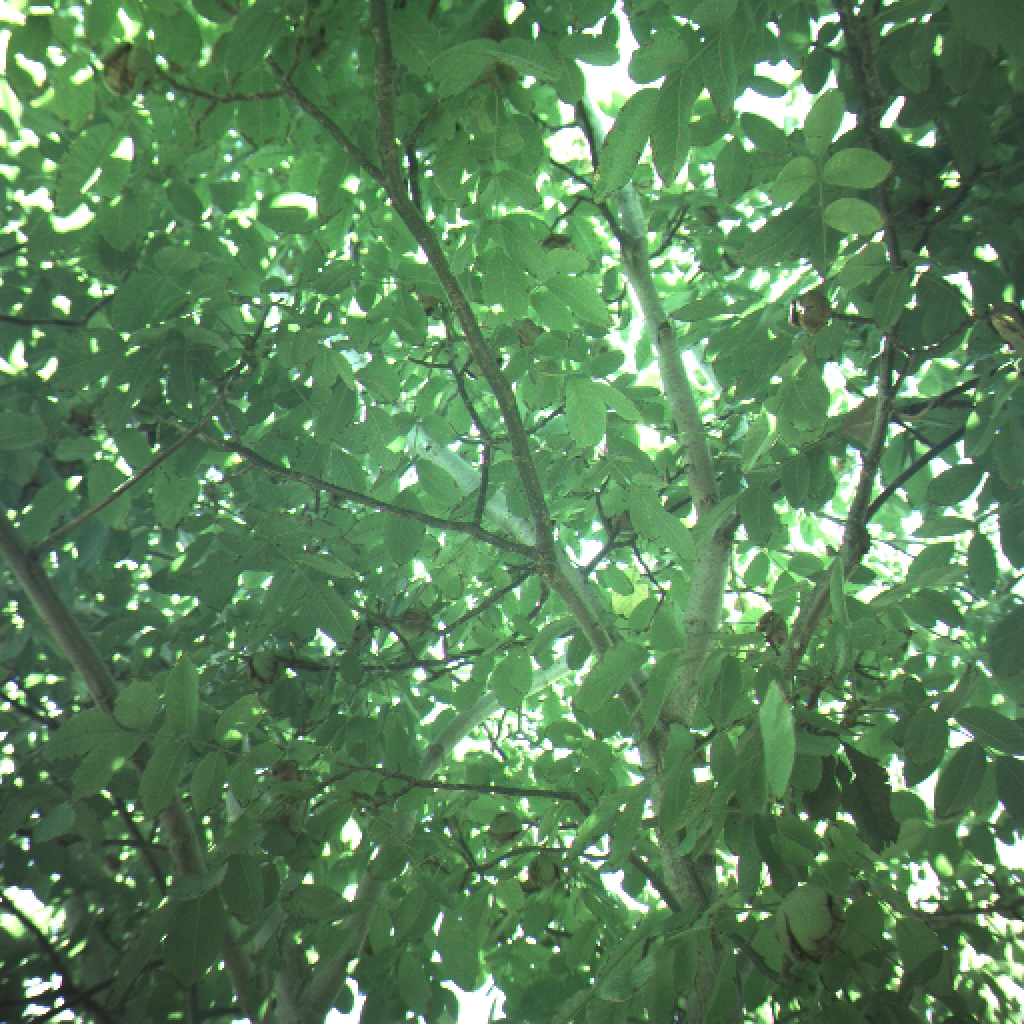}
	}
	\subfigure[]{
		\includegraphics[width=1.5in]{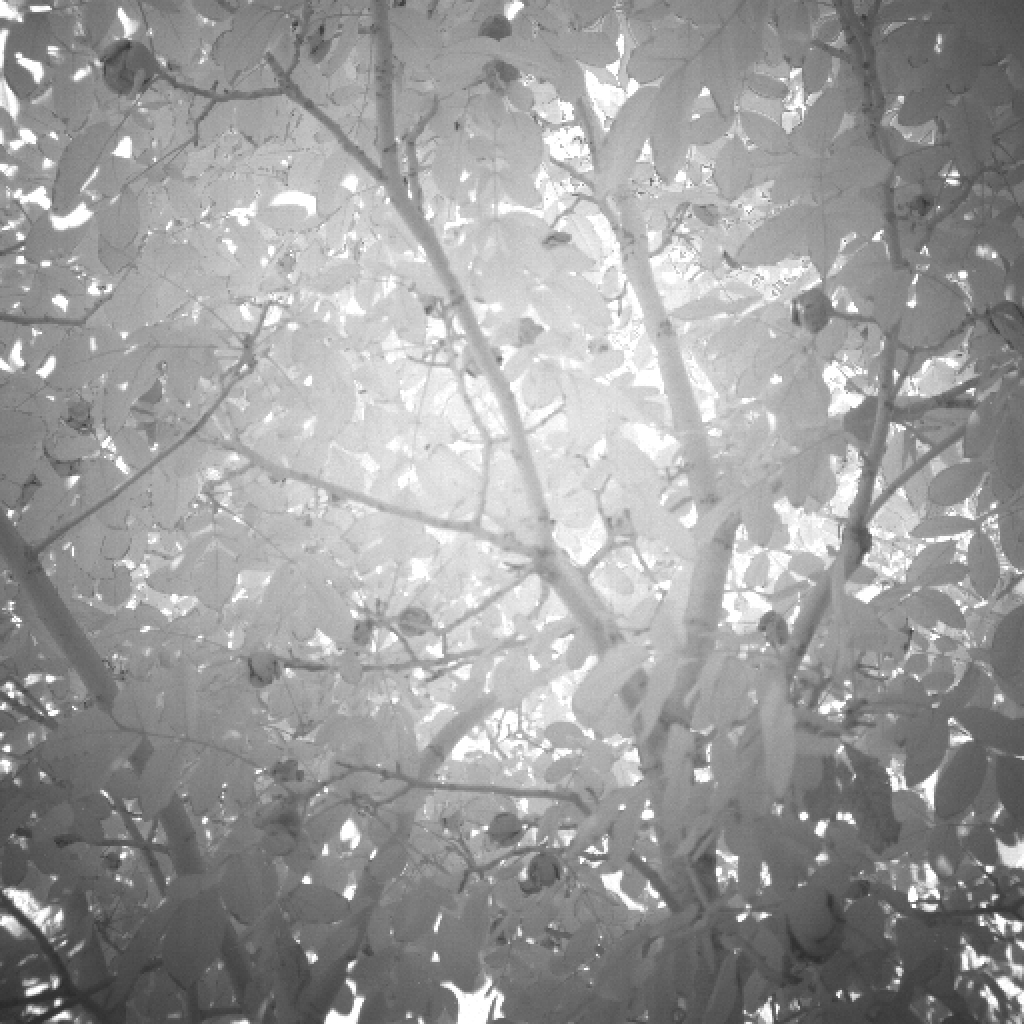}
	}	
	
	\caption{
		(a) RGB image captured using the SNAPSHOT multispectral camera. (b) NIR image from the same camera. NIR imagery aids in detecting walnuts that are nearly indistinguishable in RGB views. The camera was positioned between 50 cm and 2 m from the canopy's region of interest (ROI).}
	\label{fig:Multisepctral_Cam_RGB_NIR}
\end{figure}

\begin{figure}
	\centering
	\subfigure[]{
		\includegraphics[width=1.5in]{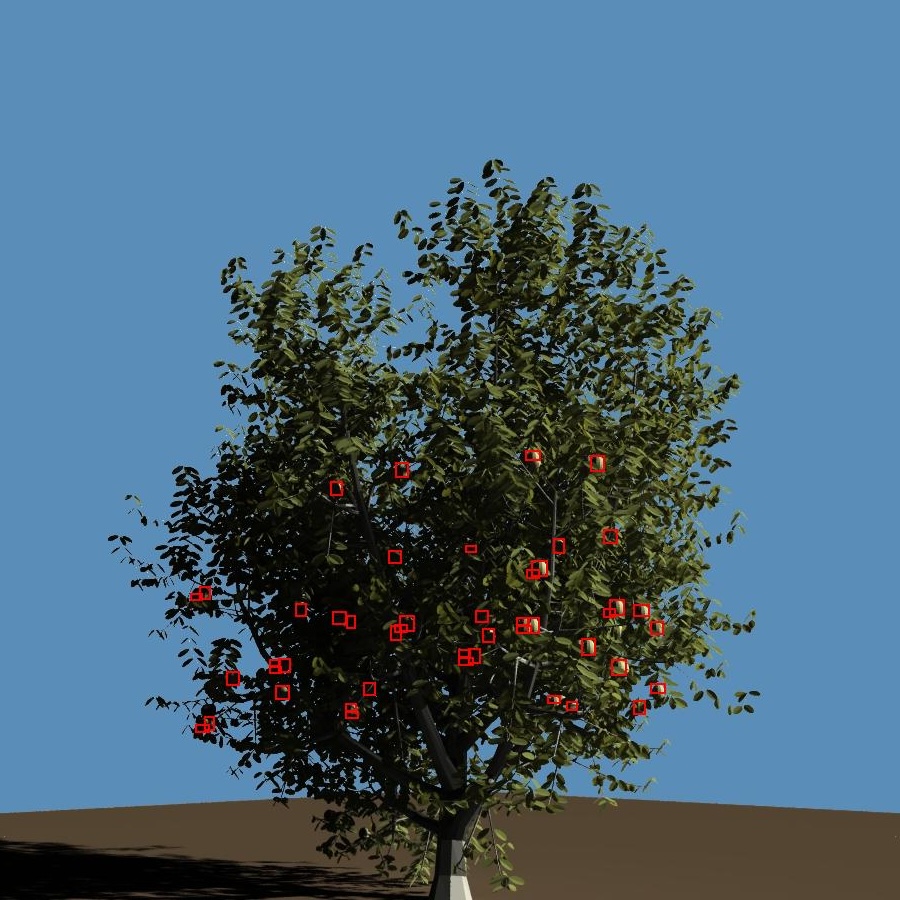}
	}
	\subfigure[]{
		\includegraphics[width=1.5in]{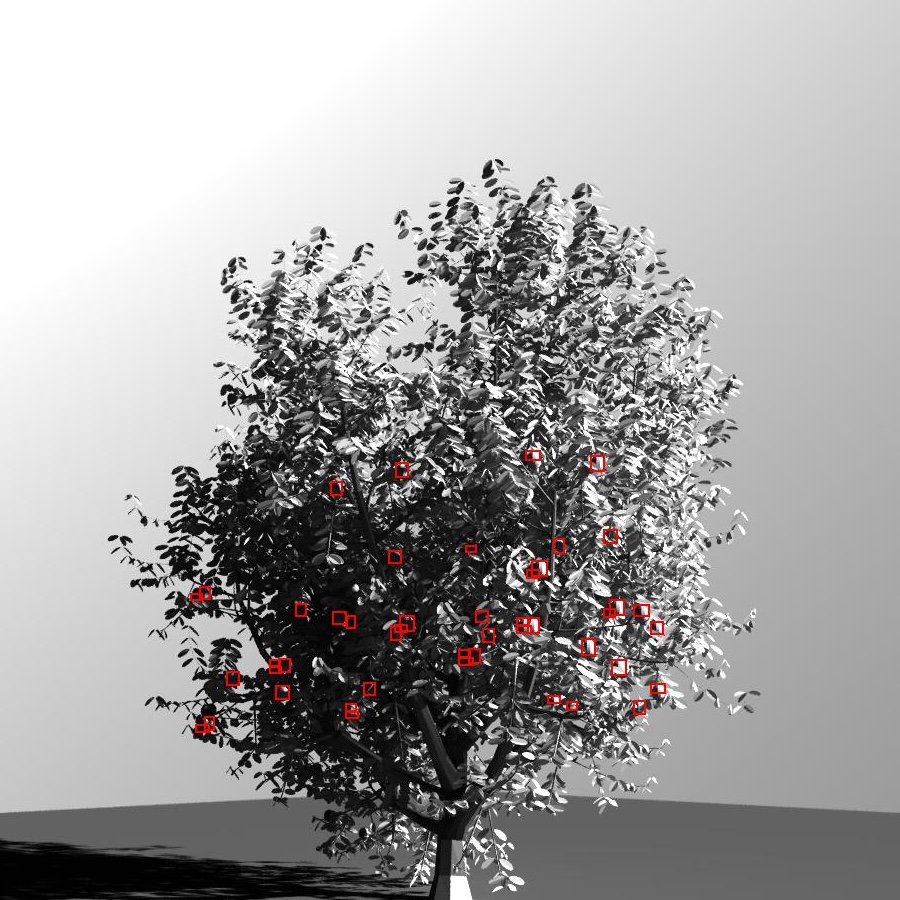}
	}	
	
	\caption{(a) Synthetic RGB image with auto-generated labels, (b) Synthetic NIR image with auto-generated labels. The labels in these synthetic images were directly produced by the simulator rather than manually annotated.}
	\label{fig:Synthetic_RGB_NIR}
\end{figure}

In September 2021, we created a set of 1500 RGB images from a Nikon COOLPIX B500 camera (Fig. \ref{fig:Nikon_RGB_NIR}). We also gathered 1500 RGB and 1500 NIR walnut images from the SNAPSHOT multispectral camera by Spectral Devices INC (Ontario, Canada) (Fig. \ref{fig:Multisepctral_Cam_RGB_NIR}). These will be used for future research on multispectral image detection (RGB + NIR). We added 500 synthetic RGB and 500 NIR walnut images to our collection (Fig. \ref{fig:Synthetic_RGB_NIR}). Both original and enhanced images were split in a consistent 4:1 ratio. For walnut detection, we used YOLOv5 and measured its precision, recall, average precision (AP), and F1 score. The NIR model was trained using three identical channels based on YOLOv5's standard setup. We enhanced the original image set with specific augmentations, and the preliminary results can be found in Table \ref{table:RGB} and Table \ref{table:NIR}.

\begin{table}[h] 
\centering
\caption{Results of original RGB image set from Nikon COOLPIX B500 camera and enhanced RGB image set with synthetic images.}
\label{table:RGB}\sffamily
\begin{tabular}{ l@{\qquad} r @{\qquad} r @{\qquad} r @{\qquad} r}
\addlinespace
\toprule
\toprule
\thead{Test Set} & \thead{Precision\\ (\%)} &  
\thead{Recall\\ (\%)} & 
\thead{AP\\ (\%)} & 
\thead{F1\\ (\%)} \\ 
\midrule
Original Image Set & 80.91 &  64.89 & 73.89 & 72.31 \\
Enhanced Image Set & 86.73 &  74.94 & 82.68 & 80.56 \\
\bottomrule
\normalsize
\end{tabular} 
\end{table}

\begin{table}[h] 
\centering
\caption{Results of original NIR image set from SNAPSHOT multispectral camera and enhanced NIR image set with synthetic images.}
\label{table:NIR}\sffamily
\begin{tabular}{ l@{\qquad} r @{\qquad} r @{\qquad} r @{\qquad} r}
\addlinespace
\toprule
\toprule
\thead{Test Set} & \thead{Precision\\ (\%)} &  
\thead{Recall\\ (\%)} & 
\thead{AP\\ (\%)} & 
\thead{F1\\ (\%)} \\ 
\midrule
Original Image Set & 78.26 &  57.66 & 68.17 & 62.07 \\
Enhanced Image Set & 85.59 &  70.32  & 78.63 & 74.48 \\
\bottomrule
\normalsize
\end{tabular} 
\end{table}

\subsection{Future Work}
This study presents our preliminary findings, emphasizing the improved walnut detection accomplished using synthetic RGB and NIR images separately. Our upcoming efforts are dedicated to exploring the capabilities of a unified multispectral model that harnesses the advantages of both RGB and NIR channels. This endeavor aims to further enhance the precision of object detection. Additionally, a promising avenue for research involves augmenting the volume of synthetic images while progressively reducing the necessity for real image dependence.

\bibliographystyle{plain}
\bibliography{ref}

\end{document}